\documentclass[lettersize,journal]{IEEEtran}
\usepackage{amsmath,amsfonts}
\usepackage{algorithmic}
\usepackage{algorithm}
\usepackage{array}
\usepackage[compatibility=false]{caption}
\usepackage{textcomp}
\usepackage{stfloats}
\usepackage{url}
\usepackage{verbatim}
\usepackage{graphicx}
\usepackage{cite}
\usepackage{tikz}
\usetikzlibrary{positioning,arrows.meta,shapes.multipart,calc}
\usepackage{hyperref}
\hypersetup{hidelinks}
\usepackage{subcaption}
\usepackage{booktabs}
\usepackage{placeins}
\usepackage{float}
\usepackage{xcolor}
\usepackage{multirow}
\usepackage{titlesec}
\titleformat{\section}{\normalfont\bfseries\scshape\centering}{\Roman{section}.}{1em}{}

\begin{document}

\title{Smartphone-Based, Field-Deployed AI Inspection System for Power Distribution Infrastructure}

\author{
Sampath~Balaji~Masabathula, 
Srija~Vanapalli,
Gangadhar~Kanchi,
V.~A.~N.~Sreenivasa~Rao,
V.~Vijaya~Lalitha,
and~Sasank~Chilamkurthy%
\thanks{S.~B.~Masabathula and S.~Vanapalli contributed equally to this work.}%
\thanks{S.~B.~Masabathula and S.~Vanapalli are with the Eastern Power Distribution Company of Andhra Pradesh Ltd., Visakhapatnam, India, and also with the Department of Computer Science and Systems Engineering, Andhra University College of Engineering, Visakhapatnam, India 
(e-mails: sb@sampathbalaji.com; srijavrao@gmail.com).}%
\thanks{G.~Kanchi and V.~A.~N.~Sreenivasa~Rao and V.~Vijaya~Lalitha are with the Eastern Power Distribution Company of Andhra Pradesh Ltd., Visakhapatnam, India
(e-mails: aee\_it5@apeasternpower.com; ee\_it@apeasternpower.com; cgm-it@apeasternpower.com).}%
\thanks{S.~Chilamkurthy is with Von Neumann AI (JOHNAIC), Bengaluru, India
(e-mail: sasank@von-neumann.ai).}%
}
\maketitle

\begin{abstract}
Routine inspection of power distribution infrastructure remains challenging for utilities operating geographically dispersed networks, where conventional manual inspection can be difficult to perform consistently and at scale. This work presents a smartphone-based AI inspection system for assessing targeted distribution-infrastructure conditions from images captured by field personnel. The proposed workflow integrates image-quality and viewing-perspective validation with AI-assisted assessment of pole lean, crossarm and top-cleat tilt, voltage level, vegetation-clearance risk, and distribution-transformer (DTR)--ground clearance. The system integrates deep-learning models with monocular depth estimation, geometric analysis, and rule-based processing within a common multi-stage inspection workflow designed for practical field operation. The system was pilot deployed across five selected distribution feeders within the Eastern Power Distribution Company of Andhra Pradesh Ltd. (APEPDCL), where 1,222 defect instances were identified across the assessed categories and subsequently tracked through a rectification and verification workflow, with 84 defect instances recorded as rectified during the study period. Feeder-level interruption records were also examined over the study period; however, the available observations do not provide sufficient evidence to establish a reliability improvement attributable to the proposed system. A curated multi-module field-captured dataset, PD-Defect, is also publicly released to support reproducible research in distribution-infrastructure inspection. The field deployment demonstrates the practical integration of smartphone-based AI-assisted inspection into field inspection and maintenance activities, providing a mechanism for systematic identification, documentation, rectification tracking, and verification of targeted distribution-infrastructure defects.
\end{abstract}


\section{Introduction}
\IEEEPARstart{P}{ower} distribution infrastructure comprises assets such as poles, conductors, crossarms, and distribution transformers (DTRs) that require periodic inspection for reliable and safe operation. Undetected defects and unsafe conditions may lead to service interruptions, safety hazards, and unplanned maintenance. However, routine field inspection of large distribution networks remains challenging for distribution companies (DISCOMs), as visual assessment often depends on the experience and judgment of field personnel, making consistent and objective evaluation across large numbers of assets difficult.

Recent advances in computer vision have demonstrated the potential of AI-assisted inspection for power distribution infrastructure, with applications including component detection, defect identification, and condition assessment. Much of the existing literature has focused on UAV-based inspection, predominantly targeting transmission infrastructure or isolated inspection tasks. Although these approaches have demonstrated promising performance, their reliance on specialized equipment limits their practicality for routine distribution-feeder inspections.

Practical deployment within DISCOMs, however, requires considerably more than accurate defect detection. Inspection systems must reliably support multiple inspection tasks, remain robust to the variability of real-world field imagery, and integrate seamlessly into existing utility inspection workflows. These practical requirements motivate the need for an integrated AI inspection framework designed specifically for routine distribution-feeder inspection.

To address these challenges, we developed a smartphone-based AI inspection system for power distribution infrastructure. The proposed system integrates image validation and multiple AI-based inspection modules within a multi-stage inspection pipeline. It provides AI-assisted assessment of pole lean, crossarm and top-cleat tilt, vegetation-clearance risk, DTR--ground clearance, and voltage level during routine field inspections. The system has been pilot-deployed within the Eastern Power Distribution Company of Andhra Pradesh Ltd. (APEPDCL), where utility field personnel have used it to conduct pole-to-pole inspections across five live distribution feeders.

The key contributions of this work are as follows: (1) a smartphone-based AI inspection system pilot-deployed within an Indian DISCOM; (2) an integrated multi-task inspection pipeline supporting multiple inspection tasks for power distribution infrastructure; and (3) the public release of PD-Defect~\cite{pddefect}, a curated multi-module smartphone-captured dataset supporting reproducible research in AI-assisted power distribution infrastructure inspection.

The underlying premise of this study is that systematic AI-assisted inspection can facilitate the identification of distribution-infrastructure defects and their subsequent rectification, which may contribute to reducing defect-related feeder interruptions and improving distribution-system reliability. Accordingly, the field evaluation examines both the performance of individual inspection modules and the use of the complete workflow for field inspection, defect rectification, and observation of feeder interruption behavior.

\section{Related Work}

Automated inspection of power infrastructure has been studied using aerial imagery and computer-vision methods. Whitworth \textit{et al.}~\cite{whitworth2001aerial} used helicopter-mounted video for visual inspection of overhead power lines. Sampedro \textit{et al.}~\cite{sampedro2019deep} developed a deep-learning system for automatic recognition and diagnosis of electrical insulator strings. More recently, Zhang \textit{et al.}~\cite{zhang2025fully} proposed a YOLOv11-based method for small-defect detection on power distribution towers using UAV imagery, while Li \textit{et al.}~\cite{li2025composite} used UAV-acquired infrared imagery for composite-insulator overheating detection. These studies address a range of inspection tasks but rely on aerial platforms or specialized sensing modalities.

Dedicated hardware has also been investigated for infrastructure monitoring. Barua \textit{et al.}~\cite{barua2025autonomous} developed an autonomous pole-climbing robot for power-line inspection and fault detection, while Masoud and Proano~\cite{masoud2026pole} proposed an autonomous system for continuous monitoring of wooden utility poles using multimodal sensing. At the ground level, Wang \textit{et al.}~\cite{wang2024leaning} proposed a low-cost monocular-camera method for estimating utility-pole leaning posture from geometric image features without prior model training or camera calibration. These studies demonstrate alternatives to aerial inspection but remain focused on specific inspection or monitoring tasks.

Across these approaches, automated inspection has primarily been applied to individual asset, defect, or condition-assessment tasks using aerial, specialized, or monocular sensing. The combination of smartphone-based image acquisition, multiple distribution-infrastructure inspection tasks, image and perspective validation, and deployment within routine DISCOM field operations is less explored in the literature reviewed here. This work addresses this setting through a unified smartphone-based inspection system integrating image validation, perspective validation, defect assessment, voltage-level classification, and geometric clearance-risk assessment.

\section{Methodology}
The proposed system is a smartphone-based AI inspection pipeline for routine inspection of power distribution infrastructure. Field personnel capture images of distribution assets using smartphones, after which the system performs input-quality, perspective, and task-specific visibility validation before directing valid inputs to the corresponding inspection modules. The pipeline integrates AI-based, rule-based, and geometric analysis for pole lean, crossarm and top-cleat tilt, voltage-level classification, vegetation-clearance risk, and DTR--ground clearance assessment. This staged design prevents unsuitable inputs from being propagated to downstream inspection tasks while allowing each module to use a method appropriate to its inspection objective.

\begin{figure*}[!t]
    \centering
    \includegraphics[width=0.93\textwidth]{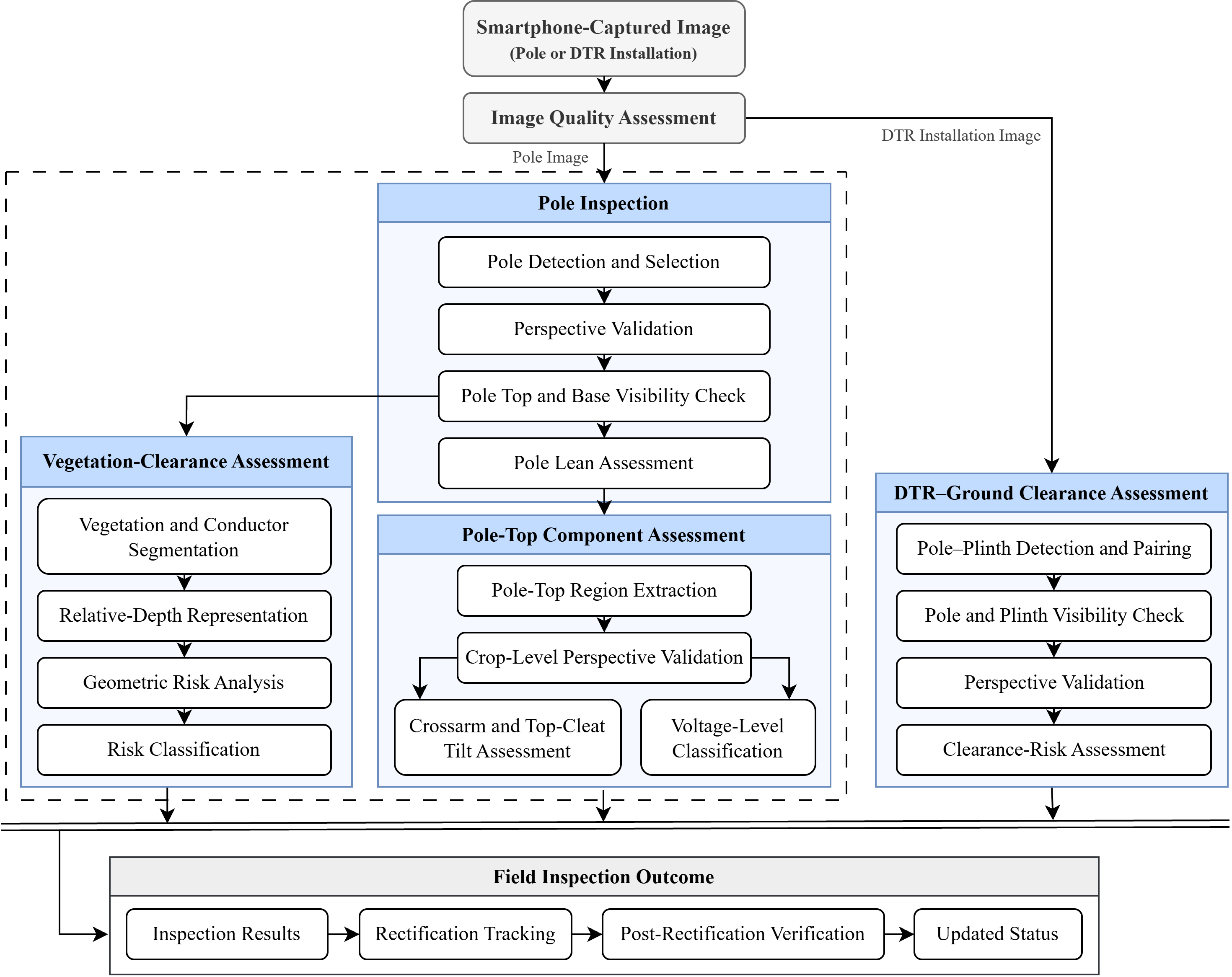}
    \caption{Overall workflow of the proposed smartphone-based AI-assisted inspection system. Smartphone-captured field images undergo input-quality assessment and are processed through the corresponding pole or DTR--ground clearance assessment pathways. Task-specific validation and assessment are then performed, and the resulting inspection outputs are recorded for rectification tracking and post-rectification verification.}
    \label{fig:overall_workflow}
\end{figure*}

The overall workflow of the proposed system is shown in Fig.~\ref{fig:overall_workflow}.

\subsection{Data Collection and Dataset Preparation}
\label{sec:data_collection}

\subsubsection{Field Data Collection}

Field images were collected by utility field personnel across urban and rural locations using entry-level Android smartphones under natural lighting conditions. The collected imagery covered distribution poles, crossarms, overhead conductors with surrounding vegetation, and DTR installations.

Initial field captures revealed that the tilt of the handheld camera could introduce angular deviations, affecting the judgment. To address this variability, the initially collected images were manually tilt-corrected under expert supervision, and a standardized capture protocol was subsequently adopted for pole and crossarm inspection. Field personnel were instructed to capture the target pole head-on from an approximately centered viewpoint, with both the pole top and base fully visible and a camera-to-pole distance of approximately 1--1.5 times the pole height. The survey application also incorporated an automatic image tilt-correction functionality adapted from the open-source Open Camera project~\cite{opencamera}, using smartphone orientation-sensor data to standardize image orientation during acquisition.

A separate capture protocol was used for DTR--ground clearance assessment. Images were required to include the pole top and base and the complete plinth on the same ground plane, with the pole and plinth visible without overlap.

\subsubsection{Dataset Organization and Filtering}

Following field acquisition, the collected images were manually curated, filtered, and organized according to the requirements of the downstream inspection tasks. The resulting data were organized into datasets for pole, crossarm, and top-cleat analysis, vegetation-clearance assessment, perspective validation, and DTR--ground clearance assessment.

For pole-lean assessment, the initial dataset comprised 2,448 smartphone-captured images. After manual quality screening, 1,801 images were retained, while 647 images were discarded because of incomplete pole visibility or perspective distortion. Following integration of automatic tilt correction into the survey application and adoption of the standardized capture protocol, an additional 2,557 images were collected. Images with incomplete pole visibility or perspective distortion that could compromise reliable assessment were filtered out, and pole-top regions of interest (ROIs) were generated from the retained imagery for crossarm, top-cleat, and voltage-level assessment. For voltage-level classification, additional pole-top crops with slight perspective distortion were included because such distortion did not materially affect voltage-level identification from the visible insulator configuration.

For vegetation-clearance risk assessment, field images were screened for visibility of both the conductor span and surrounding vegetation. Images lacking sufficient vegetation near the distribution line or adequate conductor visibility were excluded, and additional field data were collected to increase coverage of suitable vegetation--conductor scenarios. Unlike pole and crossarm assessment, this task did not require strict frontal alignment, provided that the conductor span and surrounding vegetation were captured under a reasonable viewing perspective. The resulting vegetation-clearance dataset comprised 1,781 smartphone-captured images.

A separate dataset was collected for DTR--ground clearance assessment. Of 1,148 images collected, 476 were accepted for model development and 672 were rejected because of perspective distortion or incomplete visibility of the pole or plinth bases.

Images rejected because of perspective distortion across the different inspection datasets were subsequently retained for training the DINOv3~\cite{simeoni2025dinov3} perspective-validation models. Rejected full images were used for full-image perspective validation, rejected pole-top crops were used for crop-level perspective validation, and rejected transformer images were used for perspective validation within the clearance-risk pipeline. This allowed otherwise unsuitable field images to contribute to input-quality assessment rather than being discarded entirely.

\subsubsection{Annotation Process}
\label{sec:annotation_process}

A structured annotation protocol was established to ensure consistency across the different inspection tasks. Object-detection and pose datasets were prepared in the YOLO format, while vegetation semantic segmentation annotations were prepared in the COCO format~\cite{lin2015microsoftcococommonobjects}. Domain experts defined the defect categories, visual criteria, and structural boundaries before annotation, and all annotated datasets were manually reviewed for annotation quality.

For pole-lean assessment, bounding boxes tightly enclosed the visible pole shaft from top to base, excluding crossarms, top-cleats, conductors, and background. Poles were classified as \textit{Leaned} when they exhibited a consistent visual deviation from vertical relative to the ground line and other fixed reference structures visible in the image, and as \textit{Straight} when no observable deviation from vertical was present. Only poles with clearly identifiable top and base were annotated. The corresponding pose dataset used two keypoints along the approximate pole centerline, at the top and bottom of the visible pole. When partial occlusion by vegetation or other objects occurred, the centerline was continued through the occluded region only when its geometry could be reliably inferred; otherwise, samples with indeterminate keypoints were excluded.

For crossarm and top-cleat assessment, the V-crossarm was treated as a separate class from the other crossarms because of its distinct visual structure. Bounding boxes tightly enclosed the main horizontal structural member of the crossarm, the complete V-crossarm assembly from the apex to the outer ends of both crossarm members, and the top-cleat from the upper surface to the lower clamping end. Insulators, conductors, and field-fabricated or welded extensions were excluded. Components were classified as \textit{Straight} when visually aligned with the expected structural orientation and as \textit{Tilted} when they exhibited a consistent and visually discernible deviation from the expected orientation, assessed relative to the pole axis. For partially occluded top-cleat instances, the component extent was estimated conservatively from the visible geometry.

The corresponding pose dataset used two keypoints for each pole, crossarm, V-crossarm, and top-cleat instance. For crossarms, the keypoints were placed at the centers of the two outer ends of the visible crossarm structure. For V-crossarms, the keypoints were placed at the centers of the two outer structural joints of the V-crossarm assembly. For top-cleats, the keypoints were placed along the vertical centerline at the top and bottom of the visible cleat. When the bottom clamp was clearly visible, the lower keypoint was placed at the bottom of the visible clamp. Keypoints were not extended into regions that were not visible. Partially visible or bent components were annotated only when the required keypoints could be identified reliably; otherwise, the instances were excluded from the pose dataset.

For voltage-level classification, voltage labels were assigned based on the visible insulator configuration. Crossarm and V-crossarm bounding boxes included their associated insulators, as the insulator configuration provides the basis for voltage-level classification.

Vegetation-clearance images were annotated using COCO semantic segmentation with two classes: vegetation and conductor. Vegetation was annotated only within the conductor corridor or its immediate vicinity relevant to clearance assessment; ground vegetation, distant trees, and foliage clearly separated from the conductor span were excluded. Conductor masks were annotated along the visible span from the current pole toward the adjacent pole until occlusion or the next pole. Disconnected visible conductor segments caused by occlusion were annotated as separate masks, while minor interruptions were retained within a single mask when the conductor path remained clearly identifiable.

For DTR--ground clearance assessment, poles and plinths were annotated using tight bounding boxes. Pole boxes extended from the pole top to the ground-contact point, while plinth boxes extended from the upper surface to the base. When the pole base was partially occluded, the bounding box was extended to the estimated ground-contact point.

The PD-Defect dataset used in this work is publicly available~\cite{pddefect}.

\subsection{Input Quality and Perspective Validation}

\subsubsection{Image Quality Assessment}
Each input image is first screened using a rule-based quality check based on image resolution, sharpness, pixelation, lighting, and noise. Images that do not satisfy these predefined quality criteria are discarded before downstream analysis.

\subsubsection{Pole Detection and Selection}
A YOLO12m-based detector is used to localize distribution poles from the top to the base~\cite{tian2025yolov12}. When multiple poles are detected, a rule-based selection procedure identifies the primary target pole based on its proximity to the image center and its apparent pole height. For each detected pole, a selection score is computed as
\[
S = \frac{w_1}{d_c} + w_2 h_p,
\]
where \(d_c\) is the Euclidean distance between the pole bounding-box center and the image center, \(h_p\) is the apparent pole height, and \(w_1\) and \(w_2\) are empirically selected weights used to balance the two terms. The pole with the highest score is selected for subsequent analysis. The selected pole is then subjected to a visibility check to ensure that both its top and base are sufficiently visible; poles failing this check are discarded.

\subsubsection{Full-Image Perspective Validation}
A DINOv3 ViT-B/16 model is used to validate the perspective of the full input image and reject captures with perspective distortion. The model performs binary classification between valid and perspective-distorted images. Only images classified as valid are passed to downstream inspection modules.

\subsubsection{Pole-Top Crop Validation}
A second DINOv3 ViT-B/16 model is used to validate the perspective of the extracted pole-top regions before crossarm, top-cleat assessment, and voltage-level classification. Unlike full-image validation, which evaluates the overall perspective of the input image, crop-level validation specifically assesses the local viewing geometry of the pole-top region. This additional validation is particularly important for crossarm and top-cleat assessment, where local perspective distortion can alter their apparent orientation. The same validated crops are then used for voltage-level classification to maintain consistent input conditions. Only crops classified as valid are passed to the corresponding inspection modules.

\subsubsection{DTR--Ground Clearance Image Validation}
\label{sec:DTR--Ground Clearance Image Validation}
A third DINOv3 ViT-B/16 model is used to validate the input images for DTR--ground clearance assessment. Before perspective validation, a rule-based visibility check ensures that the pole top, pole base, and plinth base are sufficiently visible. The DINOv3 model then performs binary classification between valid and perspective-distorted images to reject inputs that do not satisfy the required viewing geometry for ratio-based clearance assessment. Only images that pass both checks are passed to the clearance-risk assessment module.

\subsection{Distribution-Asset Inspection}

\subsubsection{Pole Lean Assessment}
The selected distribution pole is classified as either leaned or straight using a YOLO12m-based object-detection model. The predicted class is associated with the pre-selected target pole using intersection-over-union (IoU) matching. When multiple pole detections are present, the prediction with the highest IoU with the selected pole is retained, provided that the IoU exceeds 0.5.

To provide an interpretable measure of apparent pole inclination, a supplementary rule-based module estimates the lean angle from the pole's image-space geometry. Let $(x_1,y_1)$ and $(x_2,y_2)$ denote the two keypoints defining the detected pole axis, corresponding to the top and bottom points of the
pole, respectively. The apparent lean angle in image space, expressed in degrees, is computed as
\begin{equation}
\theta_{\mathrm{deg}} =
\frac{180}{\pi}
\operatorname{atan2}
\left(
x_1-x_2,\,
y_2-y_1
\right),
\end{equation}
where the reversed $y$-difference accounts for the downward-increasing $y$-coordinate in image space. This quantity represents an image-space measure of apparent inclination and does not correspond directly to the physical lean angle of the pole. In the deployed system, a pole is considered leaned when the estimated apparent inclination exceeds $10^\circ$.

\subsubsection{Pole-Top Region Extraction}
Following pole lean assessment, a standardized pole-top region is extracted from the selected pole for subsequent crossarm, top-cleat, and voltage-level assessment. The crop is centered on the selected pole and covers the upper 50\% of its detected height, with an additional 20\% extension upward to include the crossarm and top-cleat structures. Horizontally, the crop extends 0.57 times the detected pole height on either side of the pole center.

\subsubsection{Crossarm and Top-Cleat Tilt Assessment}
The crossarm, V-crossarm, and top-cleat detection and pose models operate on the corresponding pole-top region crop. A YOLO12m-based object-detection model predicts the orientation of the detected components as either straight or tilted. The model distinguishes three component types: Crossarm, V-crossarm, and Top-cleat.

To provide an interpretable measure of component tilt, a
supplementary rule-based module estimates the deviation of each component
from its expected mounting orientation relative to the pole. The pole,
crossarm, V-crossarm, and top-cleat axes are obtained from the
corresponding pose-model keypoints. For an axis defined by two keypoints $(x_1,y_1)$ and $(x_2,y_2)$, its
image-space orientation in degrees is computed as
\begin{equation}
\alpha =
\frac{180}{\pi}
\operatorname{atan2}
\left(
y_1-y_2,\,
x_2-x_1
\right),
\end{equation}
where the reversed $y$-difference accounts for the downward-increasing $y$-coordinate in image space. Let $\alpha_p$ and $\alpha_o$ denote the orientations of the pole and the corresponding component, respectively. Their angular separation is first computed as
\begin{equation}
\delta =
\min\left(
|\alpha_o-\alpha_p|,
360^\circ-|\alpha_o-\alpha_p|
\right).
\end{equation}
Because the pole and component are represented by undirected axes, the angular separation is then folded to the smaller angle between the two axes:
\begin{equation}
\delta_{\mathrm{acute}} =
\min\left(
\delta,180^\circ-\delta
\right),
\end{equation}
yielding an angle in the range $[0^\circ,90^\circ]$. The component tilt is subsequently defined as the deviation of this measured component--pole angle from its expected mounting angle:
\begin{equation}
\theta_{\mathrm{tilt}} =
\left|
\theta_{\mathrm{nominal}}-\delta_{\mathrm{acute}}
\right|,
\end{equation}
where $\theta_{\mathrm{nominal}}=90^\circ$ for crossarms and V-crossarms, which are expected to be perpendicular to the pole, and $\theta_{\mathrm{nominal}}=0^\circ$ for top cleats, which are expected to be aligned with the pole. Thus, a tilt of $0^\circ$ indicates the expected mounting orientation, while larger values indicate increasing angular deviation from that orientation. In the deployed system, a component is considered tilted when the computed $\theta_{\mathrm{tilt}}$ exceeds $15^\circ$.

\subsubsection{Voltage-Level Classification}
A YOLO12m-based classifier is used to predict the voltage level of the distribution line from visual cues in the pole-top region, including the insulators associated with the crossarm. The model predicts two voltage classes: High-Tension (HT) and Low-Tension (LT). The predicted voltage labels are mapped to the corresponding crossarms using a two-stage geometric matching procedure based on center-in-box and intersection-over-union (IoU) criteria. Top-cleat instances are assigned the voltage level of the nearest V-crossarm.

\subsubsection{Vegetation-Clearance Risk Assessment}

\paragraph{Problem Formulation}
Reliable assessment of vegetation-clearance risk from field-acquired smartphone images is challenging because vegetation and conductors at different depths may appear adjacent or spatially separated due to perspective projection, making pixel-space proximity an unreliable indicator of their actual spatial relationship. Consequently, semantic segmentation alone does not provide sufficient depth information to assess the spatial relationship between vegetation and conductors. The proposed method therefore combines semantic segmentation, monocular relative depth estimation, and geometric reasoning to distinguish potential clearance risks from perspective-induced false positives.

\paragraph{Vegetation and Conductor Segmentation}
Vegetation and overhead conductors are identified using a semantic segmentation model based on Mask2Former~\cite{Cheng_2022_CVPR} with a Swin-Base backbone. The resulting binary vegetation and conductor masks are denoted by $M_v$ and $M_c$, respectively. The vegetation mask is dilated using a $5\times5$ square structuring element, while the conductor mask is retained without dilation. The resulting masks are used for subsequent depth and geometric risk analysis.

\paragraph{Relative Depth Representation}
Depth Anything V2 Large~\cite{depth_anything_v2} is used to generate a monocular depth map for each input image. The resulting values represent relative rather than metric depth and are not interpreted as physical distances. Let $D(x,y)$ denote the raw depth value at pixel $(x,y)$. For each image, the valid depth values are independently min--max normalized to the range $0$--$100$, producing the normalized depth representation $Z(x,y)$. Under this depth convention, larger values of $Z(x,y)$ indicate regions that are relatively nearer to the camera, whereas smaller values indicate regions that are relatively farther away. The normalized depth map is then combined with the segmentation masks for the geometric analysis.

\paragraph{Geometric Risk Analysis}

The spatial relationship between vegetation and conductors is analyzed separately for overlapping and non-overlapping masks. Both configurations are considered potentially indicative of clearance risk, but for different reasons. In the overlap case, mask overlap does not necessarily imply physical contact, since the vegetation and conductor may lie at different depths. In the non-overlap case, spatial separation between the masks does not necessarily imply adequate clearance, since vegetation may still be sufficiently close to the conductor to pose a potential clearance risk. Accordingly, separate geometric analyses are applied to the two cases. The operating thresholds used in the geometric analysis were empirically determined on the labeled validation dataset.

The reference values used for depth-based image-scale normalization and candidate screening were empirically derived from representative field images. For each sampled image, a reference object was selected, and a tight bounding box was defined around the visible object. The normalized relative-depth value within the bounding box and the corresponding image-plane pixel length of the object were recorded for each image and aggregated across multiple images to define the reference values. This procedure yielded $Z_{\mathrm{ref}}=28.38$ on the normalized relative-depth scale and an image-plane screening threshold of 420 pixels. Because the depth estimates are relative rather than metric, these values provide a heuristic reference for image-scale normalization and candidate screening rather than a physical depth-to-distance calibration.

\textbf{1. Overlap Analysis:}
For overlapping masks, the overlap region $M_o$ is defined as the intersection of the vegetation and conductor masks. The median depth $Z_o$ of the overlap region is computed. Representative vegetation and conductor depths, $Z_v$ and $Z_c$, are obtained from a local region surrounding the conductor, with global median over the corresponding mask used when sufficient local depth information is unavailable. The depth-separation measure is defined as
\begin{equation}
\Delta_o =
\min\left(
\left|Z_o-Z_c\right|,
\left|Z_o-Z_v\right|
\right).
\end{equation}
A smaller $\Delta_o$ indicates greater depth similarity between the overlap region and at least one of the two reference regions. Based on empirically determined thresholds, an overlap risk state of high, medium, or low is assigned when $\Delta_o<9.9$, $9.9\leq\Delta_o<49.4$, or $\Delta_o\geq49.4$, respectively.

\textbf{2. Non-Overlap Analysis:}
For non-overlapping masks, candidate pairs of conductor and vegetation boundary points are evaluated as potential interaction locations. For each candidate pair $(p_c,p_v)$, the 2D Euclidean distance $x_i$ is computed. To account heuristically for variation in apparent image scale associated with the relative-depth representation, the image-plane separation $x_i$ for each candidate pair is scaled using the corresponding relative-depth value and an empirically selected reference value as
\begin{equation}
x_{\mathrm{ref}} =
x_i \frac{Z_i}{Z_{\mathrm{ref}}},
\end{equation}
where $Z_i$ is the normalized relative-depth value at the vegetation boundary point and $Z_{\mathrm{ref}}=28.38$ is an empirically selected reference value on the normalized relative-depth scale. The resulting $x_{\mathrm{ref}}$ is treated as a depth-scaled image-plane separation for candidate screening and does not represent a metric distance or a physical projection. Candidate pairs satisfying $x_{\mathrm{ref}} \leq 420$ are retained for subsequent depth-consistency analysis.

For each retained pair, the local normalized depths of the conductor and vegetation are estimated as the median of valid depth values within a $3\times3$ neighborhood of the corresponding candidate point, restricted to the respective conductor or vegetation mask. Let $Z_c$ and $Z_v$ denote the resulting median normalized depths for the conductor and vegetation, respectively. The normalized depth separation between the conductor and vegetation is then defined as
\begin{equation}
\delta = \left|Z_c-Z_v\right|.
\end{equation}
A normalized depth-separation threshold of 6 is used to identify potential vegetation--conductor interactions. If $\delta<6$ for any retained candidate pair, the non-overlap analysis indicates a potential interaction. When vegetation is present in the local near-conductor band and at least one retained pair satisfies $\delta<6$, an initial high-risk state is assigned. The minimum image-plane edge-to-edge distance between the conductor and vegetation is then evaluated. If this distance exceeds 15 pixels, the risk state is downgraded from high to low. The resulting geometric features and risk state are incorporated into a feature vector.

\paragraph{Risk Score Calibration and Operating-Point Selection}
The resulting feature vector is used by a logistic regression model to obtain a vegetation-clearance risk score. Five-fold stratified cross-validation is used to generate out-of-fold risk scores on the labeled validation dataset. An operating threshold $\tau_{p85}$ is then
selected from the precision--recall curve by maximizing recall among thresholds achieving at least 0.85 precision for the risky class. The final logistic regression model is subsequently fitted using the full labeled validation dataset, and the selected threshold is applied to its predicted risk score during deployment. An image is classified as \textbf{RISKY} when the score is greater than or equal to $\tau_{p85}$, and \textbf{SAFE} otherwise.

\subsubsection{Field-Based Module Evaluation}

For vegetation-clearance risk assessment, an independent test set was formed from available production field data collected during the pilot deployment. The reference risk labels were provided primarily by field personnel and subsequently reviewed by domain experts. The resulting test set comprised 3,298 images, including 671 \textit{Risky} and 2,627 \textit{Safe} images, and was used to evaluate the complete vegetation-clearance risk-assessment module under field conditions. These production test images were held out from the publicly released PD-Defect dataset. Although the labels were reviewed by domain experts, they are based on field assessments and may involve some degree of subjective judgment.

The validation and production test sets yielded false-positive rates of 11.02\% and 8.72\%, respectively, while the corresponding false-negative rates were 21.84\% and 6.56\%.

\subsubsection{DTR--Ground Clearance Assessment}

\paragraph{Pole--Plinth Detection and Pair Selection}
A YOLO12m-based detector is used to localize distribution poles and DTR plinths in the input image. The detected pole and plinth instances are subsequently processed to select the corresponding pole--plinth pair. The bounding-box heights of the selected pole and plinth are then used for clearance-ratio computation.

\paragraph{Clearance-Ratio Formulation}
The clearance assessment is based on the relative height of the pole and transformer plinth. Let $H$ and $h$ denote the ideal heights of the pole and plinth, respectively, and let $x$ denote an effective loss in ground clearance, such as an increase in ground level due to resurfacing. Under this change in the ground level, the effective heights of both the pole and plinth are reduced by $x$. The baseline pole-to-plinth height ratio is given by

\begin{equation}
r = \frac{H}{h}.
\end{equation}

After a clearance loss of $x$, the corresponding ratio becomes

\begin{equation}
r' = \frac{H-x}{h-x}.
\end{equation}

For $h < H$ and $0 < x < h$, $r' > r$, indicating that a reduction in effective ground clearance results in an increased pole-to-plinth height ratio. This occurs because the same clearance loss $x$ represents a larger fraction of the plinth height than of the pole height, causing the denominator to decrease proportionally more than the numerator.

Let $H_p$ and $H_{pl}$ denote the heights of the selected pole and plinth bounding boxes, respectively. The operational clearance ratio is then defined as

\begin{equation}
R = \frac{H_p}{H_{pl}}.
\end{equation}

The operational ratio $R$ therefore provides an image-based indicator of relative clearance condition and serves as an approximation to the theoretical ratio \(r'\) for inputs satisfying the required visibility and perspective conditions.

Although the current implementation is evaluated for plinth-mounted distribution transformers, the same clearance-ratio formulation can be extended to other transformer mounting configurations, including pillar-mounted and pole-mounted transformers. For these configurations, the pole and distribution transformer can be annotated using bounding boxes. Assuming that both share a common ground plane, the mounting height can be estimated from the vertical pixel distance between the bottom edge of the transformer bounding box and a horizontal baseline aligned with the bottom edge of the pole bounding box. This estimated mounting height can then replace the plinth height in the clearance-ratio formulation, allowing the same ratio-based approach to be applied to other mounting configurations. The present study is limited to plinth-mounted transformers because sufficient field data for the other mounting configurations were not available for annotation and evaluation.

\paragraph{Risk Assessment}
For valid inputs satisfying the visibility and perspective requirements described in Section~\ref{sec:DTR--Ground Clearance Image Validation}, the operational ratio $R$ is evaluated using predefined decision thresholds based on the recorded pole height $H$, in metres, obtained from the utility asset database. Assuming that one-sixth of the pole height is embedded below ground, the corresponding above-ground pole-height factor is $5/6 \approx 0.83$. The denominator values represent the baseline plinth heights used for risk classification, with $1.83$\,m representing the standard plinth height and $1.52$\,m representing the lower baseline height associated with the very-risky condition. Accordingly, the risky threshold is defined as
\begin{equation}
T_{\mathrm{risky}} = \frac{H(0.83)}{1.83},
\end{equation}
while the very-risky threshold is defined as
\begin{equation}
T_{\mathrm{very\text{-}risky}} = \frac{H(0.83)}{1.52}.
\end{equation}

The installation is classified as \textit{very risky} when
$R > T_{\mathrm{very\text{-}risky}}$, as \textit{risky} when
$T_{\mathrm{risky}} < R \leq T_{\mathrm{very\text{-}risky}}$, and as
\textit{no risk} otherwise.

\subsection{Model Training and Evaluation}

\subsubsection{YOLO12m Object-Detection Models}

YOLO12m object-detection models were used for pole localization and for multiple inspection tasks, including pole-lean assessment, crossarm, V-crossarm, and top-cleat tilt assessment, voltage-level classification, and DTR plinth--pole detection. Each model was trained on a task-specific annotated dataset.

For pole detection, a YOLO12m model was trained on 2,252 images, with 1,665 for training (73.93\%), 293 for validation (13.01\%), and 294 for testing (13.06\%). For pole-lean assessment, the object-detection dataset comprised 1,801 images, with 1,441 for training (80.01\%), 181 for validation (10.05\%), and 179 for testing (9.94\%). For crossarm, V-crossarm, and top-cleat tilt assessment, a YOLO12m model was trained on 1,686 images, with 1,175 for training (69.69\%), 254 for validation (15.07\%), and 257 for testing (15.24\%). The dataset contained six classes corresponding to straight and tilted crossarm, V-crossarm, and top-cleat instances. For voltage-level assessment, a YOLO12m model was trained on 1,772 images, with 1,337 for training (75.45\%), 218 for validation (12.30\%), and 217 for testing (12.25\%). For DTR plinth--pole detection, the released dataset contained 476 images, with 362 for training (76.05\%), 57 for validation (11.97\%), and 57 for testing (11.97\%). The dataset contained 1,191 annotated instances, comprising 699 pole and 492 plinth instances.

All YOLO12m models were trained using $640\times640$ input resolution for up to 350 epochs with mixed-precision. The optimizer was set to \texttt{auto}. The best-performing validation checkpoint was retained for final evaluation. During inference, task-specific confidence thresholds were applied: 0.35 for pole-lean and voltage-level models, 0.50 for crossarm, V-crossarm, and top-cleat tilt models, and 0.25 for transformer plinth--pole detection.

On the test sets, the pole detector achieved 97.08\% mAP@0.5, providing reliable localization for primary-pole selection and subsequent visibility validation. The DTR plinth--pole detector achieved 97.45\% mAP@0.5, with class-wise mAP@0.5 values of 98.92\% for plinths and 95.97\% for poles, supporting the subsequent ratio-based clearance assessment.

\subsubsection{YOLO12m Pose Models}

YOLO12m pose models were used to estimate structural keypoints for pole-lean and crossarm, V-crossarm, and top-cleat tilt assessment. The pole pose model operates on the full input image, whereas the component pose models operate on the corresponding pole-top-region crops. The pose models were trained on task-specific datasets with keypoint annotations. For pole-lean assessment, the keypoint dataset comprised 1,798 images, with 1,438 for training (79.98\%), 181 for validation (10.07\%), and 179 for testing (9.96\%). The dataset contained 2,025 annotated pole instances, with two keypoints per instance corresponding to the bottom and top of the pole. For crossarm, V-crossarm, and top-cleat tilt assessment, the keypoint dataset comprised 1,686 images, with 1,175 for training (69.69\%), 254 for validation (15.07\%), and 257 for testing (15.24\%). The dataset contained 4,951 annotated instances across pole, crossarm, V-crossarm, and top-cleat classes, with two keypoints per instance. Horizontal-flip augmentation was disabled for this pose model due to differences in keypoint ordering across the annotated component types.

The training configuration was kept consistent with the corresponding YOLO12m object-detection models, including the $640\times640$ input resolution, training epochs, optimizer configuration, and mixed-precision.

\subsubsection{DINOv3 Models}

Three DINOv3 ViT-B/16 models were used for perspective validation of full input images, pole-top crops, and DTR plinth--pole images. Each model performed binary classification between valid and perspective-distorted inputs. The full-image validator was trained on 2,169 images, with 1,605 for training (74.00\%), 282 for validation (13.00\%), and 282 for testing (13.00\%). These images comprised 1,018 valid and 1,151 perspective-distorted samples. The pole-top crop validator was trained on 1,743 images, with 1,221 for training (70.05\%), 260 for validation (14.92\%), and 262 for testing (15.03\%), comprising 945 valid and 798 perspective-distorted samples. The DTR plinth--pole validator was trained on 1,038 images, with 768 for training (73.99\%), 135 for validation (13.01\%), and 135 for testing (13.01\%), comprising 438 valid and 600 perspective-distorted samples.

All three models were initialized with pretrained weights, with the final four transformer blocks and a dropout-regularized linear classification head unfrozen for fine-tuning. All models used $448\times448$ input images and were trained for 75 epochs with a batch size of 16 using the AdamW optimizer, a learning rate of $1\times10^{-4}$, and weight decay of $1\times10^{-4}$. Training used the letterbox blur--jitter transformation with color jitter during training. The best-performing checkpoint based on validation accuracy was retained for the final evaluation.

\subsubsection{Mask2Former Segmentation Model}

A Mask2Former model with a Swin-Base backbone was fine-tuned for semantic segmentation of vegetation and overhead conductors. The model was initialized with the pretrained \texttt{facebook/mask2former-swin-base-ade-semantic} checkpoint from Hugging Face and trained to predict three classes: background, vegetation, and conductor. The dataset comprised 1,781 images, with 1,324 for training (74.34\%) and 457 for validation (25.66\%). Input images were resized to $1280\times1280$ pixels, with random horizontal and vertical flips and brightness, contrast, and sharpness adjustments applied as data augmentation.

Training used the AdamW optimizer with a learning rate of $1\times10^{-5}$, a cosine learning rate schedule, and a weight decay of $5\times10^{-2}$. The model was trained for up to 150 epochs with a per-device batch size of 1 and gradient accumulation over 8 steps and mixed-precision enabled.

The final Mask2Former checkpoint was selected based on downstream vegetation-clearance risk-assessment performance on the labeled validation set, using the operating-point criterion of maximizing recall subject to at least 0.85 precision for the RISKY class. For the selected checkpoint, the validation segmentation performance yielded an mIoU of 74.57\%, with IoU values of 87.64\% for vegetation and 40.77\% for conductors. Conductor segmentation achieved an F1-score of 57.92\%, with 69.51\% precision and 49.64\% recall. The resulting masks were used for relative-depth and geometric analysis; thus, pixel-level performance alone does not determine the final image-level risk classification.

For the asset inspection tasks producing image-level decisions, evaluation was performed at the image level using accuracy, precision, recall, and F1-score.

\section{Results and Discussion}
\label{sec:results}

The quantitative performance of the individual inspection and validation modules is presented in Tables I--III. The following results focus on the field deployment of the complete system, the defects identified under actual utility conditions, their subsequent rectification status, and the observed feeder interruption behavior.

\begin{table}[H]
\centering
\caption{Image-Level Classif. Metrics on Val. {\&} Test Sets}
\label{tab:img_cls_perf}
\renewcommand{\arraystretch}{1.15}
\resizebox{\columnwidth}{!}{%
\begin{tabular}{lccccc}
\toprule
\textbf{Class} & \textbf{Split} & \textbf{Acc} & \textbf{Prec} & \textbf{Rec} & \textbf{F1} \\
\midrule

\textbf{Pole Lean Assessment} & & & & & \\
Leaned\_Pole
& Test & 89.94 & 88.51 & 90.59 & 89.53 \\
& Val  & 93.92 & 93.70 & 97.54 & 95.58 \\
Straight\_Pole
& Test & 91.06 & 90.15 & 97.54 & 93.70 \\
& Val  & 95.58 & 93.42 & 95.95 & 94.67 \\

\midrule

\textbf{Crossarm and Top-Cleat Tilt Assessment} & & & & & \\
crossarm\_straight
& Test & 94.94 & 92.98 & 95.50 & 94.22 \\
& Val  & 94.49 & 91.74 & 95.24 & 93.46 \\
crossarm\_tilted
& Test & 93.00 & 78.67 & 96.72 & 86.76 \\
& Val  & 93.70 & 80.95 & 92.73 & 86.44 \\
v-crossarm\_straight
& Test & 94.55 & 91.43 & 95.05 & 93.20 \\
& Val  & 94.88 & 88.37 & 96.20 & 92.12 \\
v-crossarm\_tilted
& Test & 94.94 & 88.16 & 94.37 & 91.16 \\
& Val  & 94.49 & 90.67 & 90.67 & 90.67 \\
topcleat\_straight
& Test & 90.27 & 89.19 & 88.39 & 88.79 \\
& Val  & 93.70 & 93.52 & 91.82 & 92.66 \\
topcleat\_tilted
& Test & 93.39 & 83.87 & 88.14 & 85.95 \\
& Val  & 94.88 & 85.11 & 86.96 & 86.02 \\

\midrule

\textbf{Voltage-Level Classification} & & & & & \\
high\_tension
& Test & 97.70 & 98.82 & 98.24 & 98.53 \\
& Val  & 95.41 & 98.15 & 95.78 & 96.95 \\
low\_tension
& Test & 91.71 & 90.76 & 93.91 & 92.31 \\
& Val  & 92.66 & 91.53 & 94.74 & 93.10 \\

\midrule

\textbf{Vegetation-Clearance Risk Assessment} & & & & & \\
Risky
& Test & 93.44 & 73.25 & 93.44 & 82.12 \\
& Val  & 78.16 & 85.64 & 78.16 & 81.73 \\
Safe
& Test & 91.28 & 98.20 & 91.28 & 94.61 \\
& Val  & 88.98 & 82.89 & 88.98 & 85.83 \\
\bottomrule
\end{tabular}}
\end{table}

\begin{table}[H]
\centering
\caption{Perspective-Validation Metrics on Val. \& Test Sets}
\label{tab:perspective_perf}
\renewcommand{\arraystretch}{1.15}
\resizebox{\columnwidth}{!}{%
\begin{tabular}{lccccc}
\toprule
\textbf{Class} & \textbf{Split} & \textbf{Acc} & \textbf{Prec} &
\textbf{Rec} & \textbf{F1} \\
\midrule

\textbf{Full-Image} & & & & & \\
Valid
& Test & 91.67 & 88.32 & 91.67 & 89.96 \\
& Val  & 89.47 & 88.15 & 89.47 & 88.81 \\
perspective-distorted
& Test & 89.33 & 92.41 & 89.33 & 90.85 \\
& Val  & 89.26 & 90.48 & 89.26 & 89.86 \\

\midrule

\textbf{Pole-Top Crop} & & & & & \\
Valid
& Test & 97.18 & 96.50 & 97.18 & 96.84 \\
& Val  & 97.16 & 93.84 & 97.16 & 95.47 \\
perspective-distorted
& Test & 95.83 & 96.64 & 95.83 & 96.23 \\
& Val  & 92.44 & 96.49 & 92.44 & 94.42 \\

\midrule

\textbf{DTR plinth--pole} & & & & & \\
Valid
& Test & 84.21 & 81.36 & 84.21 & 82.76 \\
& Val  & 89.47 & 89.47 & 89.47 & 89.47 \\
perspective-distorted
& Test & 85.90 & 88.16 & 85.90 & 87.01 \\
& Val  & 92.31 & 92.31 & 92.31 & 92.31 \\

\bottomrule
\end{tabular}}
\end{table}

\begin{table}[H]
\centering
\caption{Pose Estimation and Derived Angular Metrics}
\label{tab:pose_perf}
\renewcommand{\arraystretch}{1.15}
\resizebox{\columnwidth}{!}{%
\begin{tabular}{lccccc}
\toprule
\textbf{Class} & \textbf{Split} & \textbf{Prec} &
\textbf{Rec} & \textbf{mAP@0.5} & \textbf{Angular MAE ($^\circ$)} \\
\midrule

\textbf{Pole (Lean)} & & & & & \\
Pole
& Test & 96.86 & 93.40 & 97.75 & \textbf{0.29} \\
& Val  & 96.47 & 90.33 & 98.19 & 0.60 \\

\midrule

\textbf{Pole-top Component} & & & & & \\
Pole
& Test & 99.24 & 98.10 & 99.39 & \textbf{0.37} \\
& Val  & 96.85 & 94.49 & 98.09 & 0.38 \\
Crossarm
& Test & 94.29 & 92.30 & 97.23 & \textbf{0.51} \\
& Val  & 93.52 & 91.71 & 95.87 & 0.43 \\
V-crossarm
& Test & 99.65 & 99.45 & 99.50 & \textbf{0.51} \\
& Val  & 98.74 & 97.01 & 99.39 & 0.52 \\
Top-cleat
& Test & 93.24 & 93.66 & 97.78 & \textbf{2.01} \\
& Val  & 89.22 & 95.52 & 96.49 & 1.67 \\

\bottomrule
\end{tabular}}
\end{table}

\begin{figure*}
    \centering
    \includegraphics[width=0.77\linewidth]{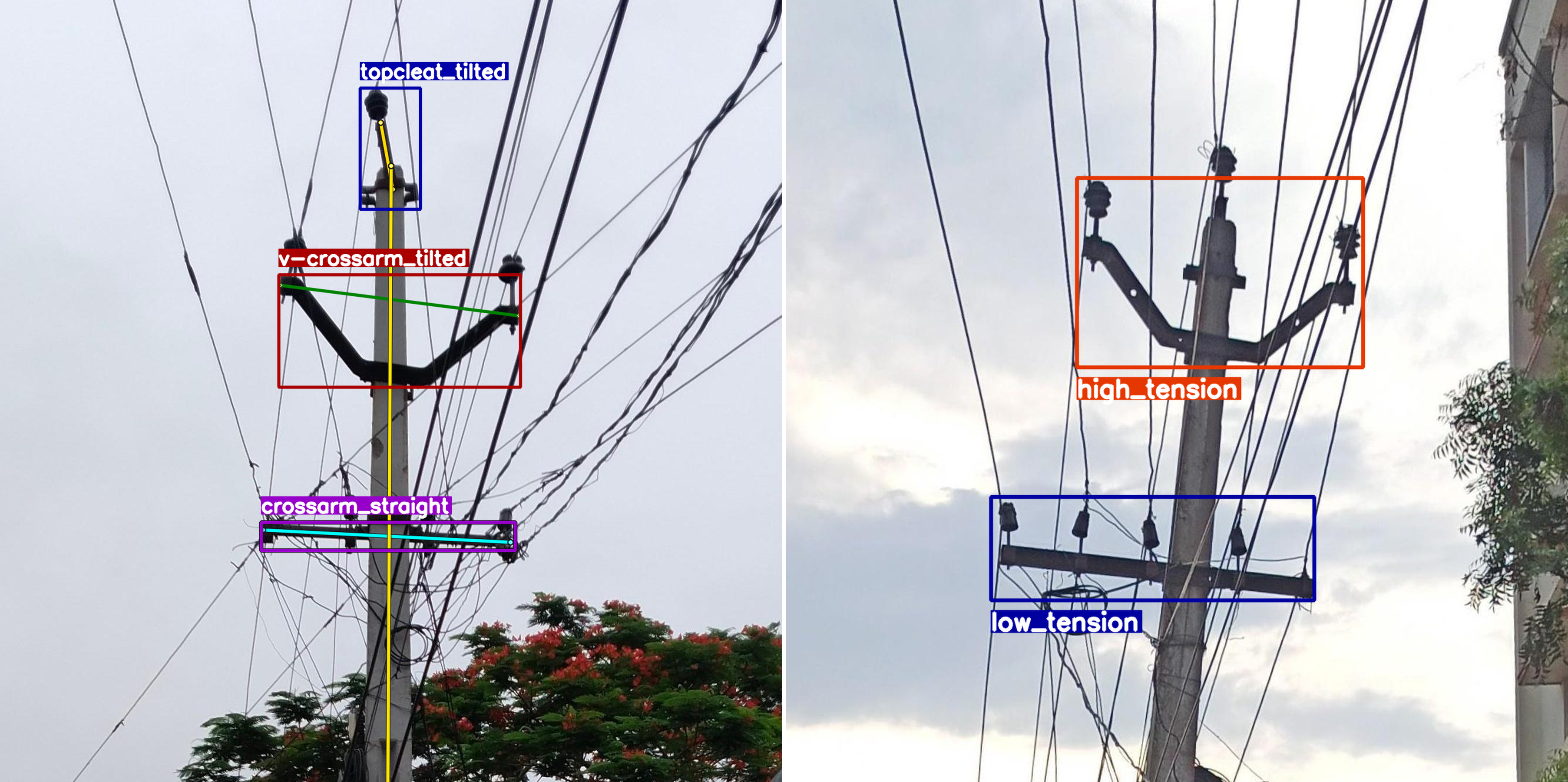}
    \caption{Representative outputs showing (a) detected crossarm, V-crossarm, and top-cleat components with tilt angles of $2.86^\circ$, $7.26^\circ$, and $10.52^\circ$, respectively, estimated from axes defined by their detected keypoints. Although tilt is detected for the V-crossarm and top-cleat, their angles remain below the $15^\circ$ threshold and are therefore classified as non-defective; and (b) detected high-tension (HT) and low-tension (LT) structures for voltage-level classification.}
    \label{fig:arm_and_voltage_image}
\end{figure*}

\begin{figure}[H]
    \centering
    \includegraphics[width=0.75\linewidth]{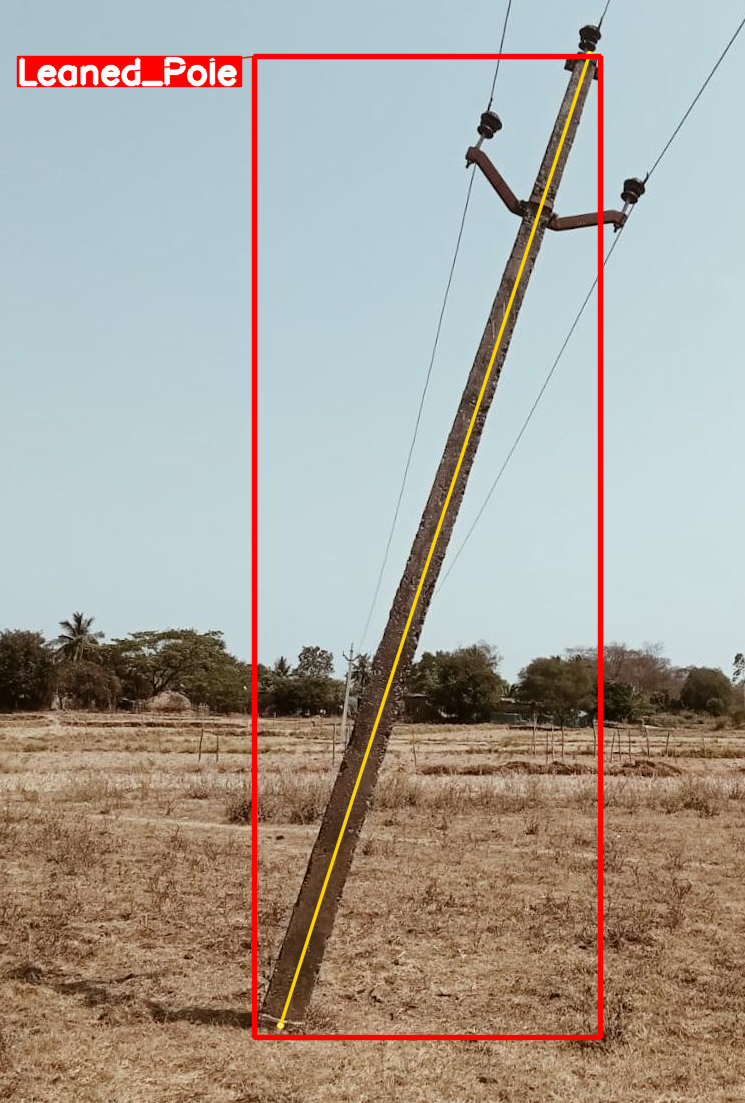}
    \caption{Representative pole-lean assessment showing a detected pole with an apparent inclination of $17.82^\circ$, estimated from the pole axis defined by its detected keypoints, resulting in a leaned classification.}
    \label{fig:pole_lean}
\end{figure}

\subsection{Field Deployment and Inspection}

To evaluate the operational utility of the proposed system beyond model-level performance, the system was pilot deployed across five selected distribution feeders under actual utility field conditions. Field personnel used the smartphone-based inspection application to capture images of distribution assets and assess the supported inspection categories during the feeder inspection process.

Across the five feeders, the inspection identified 1,222 defect instances associated with 1,141 distinct poles. Vegetation-clearance risk accounted for the largest share, with 900 instances, followed by pole-lean defects (157), top-cleat-related defects (60), crossarm-related defects (93), and DTR–ground clearance defects (12). The feeder-wise distribution of the identified defect instances is presented in Table~\ref{tab:field_defects}.

\begin{table*}[!t]
\centering
\caption{Defect Identification Across the Five Selected Feeders. The \textit{Defective} column denotes distinct poles associated with identified defects; Crossarm and Top-Cleat denote tilt-related defect instances.}
\label{tab:field_defects}
\renewcommand{\arraystretch}{1.15}
\setlength{\tabcolsep}{6pt}
\begin{tabular}{lrrrrrrrr}
\toprule
\textbf{Feeder} &
\textbf{Surveyed} &
\textbf{Defective} &
\textbf{Pole Lean} &
\textbf{Crossarm} &
\textbf{Top-Cleat} &
\textbf{DTR--Ground Clearance} &
\textbf{Vegetation Clearance} &
\textbf{Total} \\
\midrule
11KV BONDADA & 557 & 157 & 16 & 12 & 11 & 0 & 128 & 167 \\
11KV PERUMALI & 807 & 214 & 62 & 25 & 12 & 1 & 125 & 225 \\
11KV YERRAKONERU & 853 & 333 & 43 & 19 & 21 & 4 & 272 & 359 \\
11KV BADIVANIPETA & 836 & 317 & 19 & 19 & 9 & 4 & 287 & 338 \\
11KV A.RAVIVALASA & 757 & 120 & 17 & 18 & 7 & 3 & 88 & 133 \\
\midrule
\textbf{Total} & \textbf{3,810} & \textbf{1,141} & \textbf{157} &
\textbf{93} & \textbf{60} & \textbf{12} & \textbf{900} &
\textbf{1,222} \\
\bottomrule
\end{tabular}
\end{table*}

\subsection{Defect Findings and Rectification Status}

The identified defects were automatically recorded in the inspection application as pending rectifications. Following physical rectification, field personnel captured a post-rectification image of the corresponding asset through the application, and the system validated whether the identified defect had been resolved. When a defect was fully resolved, the corresponding rectification was accepted and cleared. If only part of the identified defect was resolved, the resolved portion was cleared while the remaining defect continued to be recorded as pending. Unsuccessful rectification attempts likewise remained pending for further action.

During the study period, 84 defect cases were recorded as rectified across the five selected feeders, comprising 34 vegetation-clearance instances, 32 crossarm-tilt instances, 15 top-cleat-tilt instances, and 3 pole-lean instances. Consequently, only a subset of the identified conditions could be physically rectified and subsequently verified within the study period.

\subsection{Feeder Interruption Analysis}

To examine feeder-level interruption behavior in the context of the field deployment, all recorded interruption events for the five selected feeders were analyzed over the study period. The interruption-type field was not used for filtering, as the recorded type is operator-entered and does not necessarily represent the actual cause of the interruption. Accordingly, the analysis considers the complete set of recorded interruption events without cause-based filtering. April through June 2026 were used as the baseline reference period, preceding the commencement of the field survey in July 2026. The field survey was conducted during July 2026 over approximately three weeks, followed by the subsequent observation period beginning in August 2026.

Across the five feeders, the April--June 2026 baseline period comprised 893 recorded interruption events, corresponding to a monthly average of 297.67 events. During July 2026, 273 interruption events were recorded, followed by 180 events in August and 334 events in September. The feeder-wise monthly interruption records are presented in Table~\ref{tab:feeder_interruption_analysis}.

The monthly interruption records show variation in feeder interruption behavior over the study period. The number of recorded interruption events decreased during August relative to the April--June baseline average and increased again in September. Such month-to-month variation may be influenced by multiple network and operational factors, including weather conditions, equipment failures, supply-side issues, and other events outside the scope of the present inspection workflow. However, the available interruption records do not provide sufficient information to isolate the contribution of these factors or to attribute the observed variation to the proposed inspection system. Therefore, the interruption records are interpreted as field-level reliability observations rather than as a direct measure of the effect of the proposed system on feeder reliability.

\begin{table}[!t]
\centering
\caption{Feeder-Wise Monthly Interruption Records Across the Study Period}
\label{tab:feeder_interruption_analysis}
\renewcommand{\arraystretch}{1.15}
\resizebox{\columnwidth}{!}{%
\begin{tabular}{lrrrrrr}
\toprule
\textbf{Feeder} & \textbf{Apr.} & \textbf{May} &
\textbf{Jun.} & \textbf{Jul.} & \textbf{Aug.} & \textbf{Sep.} \\
\midrule
11KV A.RAVIVALASA & 59 & 79 & 91 & 65 & 42 & 86 \\
11KV BADIVANIPETA & 54 & 40 & 52 & 29 & 30 & 63 \\
11KV BONDADA & 32 & 68 & 36 & 53 & 17 & 42 \\
11KV PERUMALI & 34 & 62 & 79 & 50 & 51 & 78 \\
11KV YERRAKONERU & 48 & 78 & 81 & 76 & 40 & 65 \\
\midrule
\textbf{Total} & \textbf{227} & \textbf{327} & \textbf{339} &
\textbf{273} & \textbf{180} & \textbf{334} \\
\bottomrule
\end{tabular}}
\end{table}

\subsection{Operational Implications and Limitations}

The field deployment demonstrates the practical integration of the proposed smartphone-based AI-assisted inspection workflow into field inspection and maintenance activities. The workflow provides a mechanism for systematic identification, documentation, rectification tracking, and verification of targeted distribution-infrastructure defects within a common inspection process.

Only a subset of the identified defects could be physically rectified and subsequently verified during the study period, primarily because of operational constraints in obtaining the line-clearance permissions required for rectification work on energized distribution infrastructure. Such clearances require temporary interruption of power supply and must therefore be coordinated with feeder operating conditions and other planned or emergent network activities.

This limitation also constrains the extent to which downstream maintenance outcomes could be evaluated. In addition, feeder interruption behavior can be influenced by multiple network and operational factors, including conditions beyond those addressed through the current inspection workflow. The available field observations therefore do not provide sufficient evidence to establish a reliability improvement attributable to the proposed system. The results instead demonstrate the operational utility of AI-assisted inspection as a mechanism for systematic defect identification, documentation, rectification tracking, and verification within a broader distribution-system maintenance approach, in which identified defects can be addressed alongside other network maintenance and operational activities.

\section{Conclusion and Future Work}
\label{sec:conclusion}

This study presented a smartphone-based, field-deployed AI inspection workflow for assessing targeted distribution-infrastructure conditions using images captured by field personnel. The workflow integrates image-quality and viewing-perspective validation with AI-assisted assessment of pole lean, crossarm, V-crossarm, and top-cleat tilt, voltage level, vegetation-clearance risk, and DTR--ground clearance within a common inspection process.

The system was pilot deployed across five selected distribution feeders under actual utility field conditions. During the deployment, the system identified 1,222 defect instances across the assessed defect categories. These defects were subsequently tracked through a rectification and verification workflow, with 84 cases recorded as rectified during the study period. The field deployment demonstrates the practical integration of the proposed smartphone-based AI-assisted inspection workflow into field inspection and maintenance activities.

Feeder-level interruption records were also examined over the study period. The observed variation in interruption behavior did not provide sufficient evidence to establish a reliability improvement attributable to the proposed system. This limitation is primarily associated with the limited number of identified defects that were physically rectified during the study period and the potential variation in feeder interruptions due to multiple network and operational factors. Consequently, the observed interruption behavior should be interpreted as a field-level reliability observation rather than as a direct measure of the effect of the proposed inspection workflow on feeder reliability.

The proposed system provides a practical foundation for incorporating AI-assisted inspection into periodic distribution-infrastructure maintenance. Future work will focus on extending deployment across a larger number of feeders and longer observation periods, increasing the number of identified defects that are physically rectified and subsequently verified, expanding the range of distribution-infrastructure defects assessed by the system, and monitoring feeder interruption behavior over longer periods. Such extended field studies can provide a stronger basis for evaluating the relationship between AI-assisted defect identification, subsequent rectification, and distribution-system reliability.

\section*{Acknowledgment}

The authors express their sincere gratitude to Prudhvi Tej Immadi, IAS, Chairman and Managing Director, and Sri T. V. Surya Prakash, Director--Operations, of the Eastern Power Distribution Company of Andhra Pradesh Ltd. (APEPDCL), for their support for this initiative. The authors also acknowledge APEPDCL, Visakhapatnam, for its support and cooperation. Special thanks are extended to the field data collection team and to Marla Krishna Sumwith and Shreya Vanapalli for their contributions to the annotation of the dataset used in this study.

\bibliographystyle{IEEEtran}
\bibliography{refs}

\end{document}